\documentclass[preprint,12pt]{elsarticle} 

\usepackage{amssymb}

\makeatletter
\newcommand{\removelatexerror}{\let\@latex@error\@gobble}
\makeatother

\usepackage{hyperref}
\usepackage{url}
\usepackage{subcaption}
\usepackage{amssymb}
\usepackage{amsmath}
\usepackage{amsthm,amsmath,amssymb}
\usepackage{mathrsfs}

\usepackage[flushleft]{threeparttable}
\usepackage{threeparttable}
\usepackage{booktabs}
\usepackage{comment}
\usepackage{amsthm}
\usepackage{amsmath}
\usepackage{apxproof}
\usepackage{thmtools, thm-restate}
\usepackage{wrapfig}
\usepackage{adjustbox}
\usepackage[flushleft]{threeparttable}
\usepackage{graphicx}

\usepackage{dashrule}

\usepackage[noamssymbols,upint]{newtxmath} 

\usepackage[ruled,linesnumbered]{algorithm2e}
\usepackage{algorithmic}
\usepackage{float}
\usepackage{ulem}
\usepackage{cancel}

\DeclareMathAlphabet\mathbfcal{OMS}{cmsy}{b}{n}

\usepackage{microtype}
\usepackage{graphicx}

\usepackage{booktabs} 

\usepackage{hyperref}
\usepackage{amssymb}
\usepackage{amsmath}
\usepackage{amsthm}

\usepackage{adjustbox}

\usepackage{cancel}

\usepackage[inline]{enumitem}

\usepackage{bm}
\usepackage{makecell}
\usepackage[figuresright]{rotating}

\usepackage{adjustbox}

\usepackage{enumitem}

\usepackage{graphicx}
\usepackage{array}

\usepackage{extarrows}
\usepackage{caption}
\usepackage{graphicx}
\usepackage{sidecap}
\usepackage{float}
\usepackage{framed}
\usepackage{tabularx}
\usepackage{twoopt}
\usepackage{adjustbox}

\usepackage{lscape}

\usepackage[table]{xcolor}

\usepackage{tikz}
\usetikzlibrary{fit}

\usepackage{booktabs,threeparttable}

\usepackage{lineno}

\newcommand{\yali}[1]{{\color{magenta}[Yali: #1]}}

\newcommandtwoopt\Textbox[5][7.2cm][2cm]{%
\begin{tikzpicture}[remember picture,overlay]
  \coordinate (aux) at ([xshift=#1]#4);
  \node[inner ysep=3pt,yshift=1ex,draw=pink,thick,
    fit=(#3) (aux),baseline] 
    (box) {};
  \node[text width=#2,anchor=north east,
    font=\sffamily\footnotesize,
  align=right
    ] 
    at (box.north east) {#5};
\end{tikzpicture}%
}

\hypersetup{
colorlinks=true,
citecolor=red,
    linkcolor=red,
    filecolor=magenta,      
    urlcolor=red,
linktocpage}

\usepackage{lipsum}


\begin{document}

\begin{frontmatter}
\title{A Human-Centered Safe Robot Reinforcement Learning
Framework with Interactive Behaviors}




\author{Shangding Gu\fnref{label1}$^{*}$%
}
\author{Alap Kshirsagar\fnref{label2}$^{*}$%
}
\author{Yali Du\fnref{label3}$^{*}$%
}
\author{Guang Chen \fnref{label4}%
}
\author{Jan Peters\fnref{label2}%
}
\author{Alois Knoll\fnref{label1}%
}
\cortext[cor1]{These authors contributed equally to this work. } 
\fntext[label1]{Technical University of Munich,  Munich, 85748, Germany.%
}
\fntext[label2]{Technical University of Darmstadt,  Darmstadt, 64289, Germany. %
}
\fntext[label3]{King’s College London, London, WC1E 6EB, UK.%
}
\fntext[label4]{Tongji University,  Shanghai, 201804, China.%
}


\begin{abstract}
Deployment of Reinforcement Learning (RL) algorithms for robotics applications in the real world requires ensuring the safety of the robot and its environment. Safe Robot RL (SRRL) is a crucial step towards achieving human-robot coexistence. In this paper, we envision a human-centered SRRL framework consisting of three stages: safe exploration, safety value alignment, and safe collaboration. We examine the research gaps in these areas and propose to leverage interactive behaviors for SRRL. Interactive behaviors enable bi-directional information transfer between humans and robots, such as conversational robot ChatGPT \cite{OpenAI_gpt}. 
We argue that interactive behaviors need further attention from the SRRL community. We discuss four open challenges related to the robustness, efficiency,  transparency, and adaptability of SRRL with interactive behaviors. 

\end{abstract}



\begin{keyword}
  Interactive Behaviors; Safe Exploration; Value Alignment; Safe Collaboration; Bi-direction Information.



\end{keyword}

\end{frontmatter}






\section{Introduction} 



\begin{figure}[htbp!]
  \centering
  \includegraphics[width=0.99\linewidth]{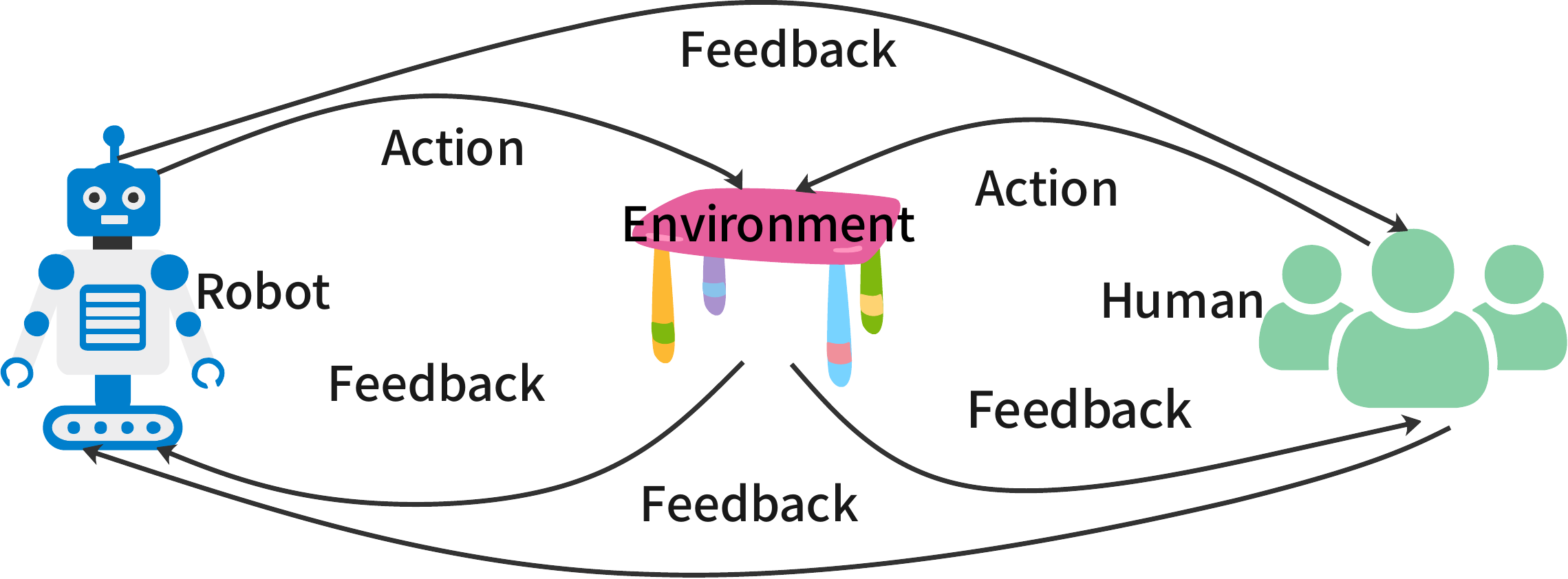}
  \caption{Schematic of interactive behaviors.}
  \label{fig:safe-robot-rl-interactive-bahvaiors}
\end{figure}

Deep learning has shown impressive performance in recent years~\cite{lecun2015deep}. By leveraging deep learning, Reinforcement Learning (RL) has achieved remarkable successes in many scenarios and superhuman performance in some challenging tasks ~\cite{gu2022review, gu2021multi}, e.g., autonomous driving~\cite{gu2022constrained}, recommender system~\cite{zhao2021dear}, robotics~\cite{brunke2021safe},  {games}~\cite{silver2018general,du2019liir,han2019grid}, and finance~\cite{tamar2012policy}. Most RL methods aim to maximize reward performance without considering safety constraints. However, safety is critical when deploying RL in real-world applications, especially in robotics. In Robot RL (RRL), a robot interacts with static or dynamic environments to learn the probability of better actions. When humans are also part of the robot's environment, ensuring their safety is crucial. This paper proposes a framework to achieve Safe Robot RL (SRRL) by leveraging interactive behaviors. 
Interactive behaviors are behaviors that can mutually influence the interacting elements. Interaction is everywhere in human life~\cite{kong2018human}, and agent-environment interaction is the basis of RL~\cite{sutton2018reinforcement}. When humans and robots act in a shared environment, their actions can be influenced by each other through interactive behaviors~\cite{macglashan2017interactive, knox2009interactively, thomaz2006reinforcement,lou2023pecan,kazantzidis2022train}. For example, a robot navigating in a public space can plan its path to avoid collisions with pedestrians and provide signals to pedestrians to move aside. The pedestrians can also plan their path to avoid collisions with the robot and provide signals to the robot to move aside. Figure~\ref{fig:safe-robot-rl-interactive-bahvaiors} shows a schematic of interactive behaviors consisting of three elements: robots, environment, and humans. The outer loop consists of feedback or signals from the robots to the humans and vice versa. The two inner loops consist of actions from robots and humans in the environment and feedback or rewards from the environment to robots and humans. Here, we use the term ``feedback'' to mean any type of information transfer between the two interacting elements.




Interactive behaviors can lead to better SRRL by enabling bi-directional information transfer between the interacting elements. It is a core technology to improve the dialogue performance of a Large Language Model (LLM), e.g., ChatGPT \cite{OpenAI_gpt}
by leveraging interactive behaviors. For example, in ChatGPT, on the one hand, high-quality data from human feedback is collected to design the reward model~\cite{gao2022scaling, stiennon2020learning}. On the other hand, after having human feedback, the agent model will be trained to align human values, and then give safe feedback to humans 
. In most conventional SRRL approaches, there is no human interaction, e.g., CPO~\cite{achiam2017constrained} and ATACOM~\cite{liu2022robot}. With interactive behaviors, the robot can learn about human behaviors and convey its features and decision-making processes to humans. Those approaches that consider the human-in-the-loop of the robot's learning process do not utilize feedback from the robot learner to the human teacher.


In this paper, we investigate interactive behaviors to achieve three stages of human-centered SRRL: ``safe exploration'', ``safety value alignment'',  and ``safe human-robot collaboration''. In the ``safe exploration'' stage, the robot must explore the unknown state space while preserving safety. In the ``safety value alignment'' stage, the robot has to align its intentions with the humans. Finally, in the ``collaboration" stage, the robot should contribute towards achieving shared goals with humans.

\section{Related Work on Safe Robot-Reinforcement Learning}


Safe robot learning has received substantial attention over the last few decades~\cite{turchetta2019safe, baumann2021gosafe, kaushik2022safeapt, marco2021robot, kroemer2021review}. SRRL methods focus on robot action and state optimization and modeling to ensure the safety of robot learning. For instance, 
Gaussian models are used to model the safe state space~\cite{akametalu2014reachability, sui2015safe, turchetta2016safe, berkenkamp2016safe, sui2018stagewise, wachi2018safe}; formal methods are leveraged to verify safe action and state space~\cite{fulton2018safe, kochdumper2022provably, yu2022safe}; control theory is applied to search safe action space~\cite{chow2018lyapunov, chow2019lyapunov, marvi2021safe, koller2018learning, li2019temporal}.

Anayo \textit{et al.} \cite{akametalu2014reachability} introduced a reachability-based method to learn system dynamics by using Gaussian models, in which the agent can adaptively learn unknown system dynamics and the maximal safe set. Berkenkamp \textit{et al.}~\cite{berkenkamp2016safe} proposed a region of attraction method to guarantee safe state space based on Gaussian processes and Bayesian optimization. Although the method provided a theoretical analysis for safe robot learning, the assumptions in the study are quite strong and may not be applicable in practical environments. Sui \textit{et al.}~\cite{sui2015safe} present a safe exploration method considering noise evaluations, in which the safety is ensured with a high likelihood via a Gaussian process confidence bound. Moreover, the sample complexity and convergence of the method are analyzed, and real-world applications, such as movie recommendations and therapeutic spinal cord stimulation, are used to test how well the system works. Nonetheless, they considered the safe exploration as a bandit setting without any constraints~\cite{turchetta2016safe}. Turchetta \textit{et al.}~\cite{turchetta2016safe} developed an algorithm in which Gaussian processes are leveraged to model the safe constraints. However, a set of starting safe states is required from which the agent can begin to explore in the work of Sui \textit{et al.}~\cite{sui2015safe} and Turchetta \textit{et al.}~\cite{turchetta2016safe}.


Some recent works have investigated the application of formal methods for SRRL. Nathan and  Andre \cite{fulton2018safe} used formal verification to check the correctness of the state transition model and select safe action during RL. They allow unsafe actions if the model is incorrect. Researchers have proposed three types of provably safe RL methods for hard safety~\cite{krasowski2022provably}: action mask, action replacement, and action projection. Action masking approaches apply a safety layer to restrict the agent's actions to safe actions only~\cite{krasowski2020safe}. Action replacement approaches replace unsafe actions with safe actions~\cite{hunt2021verifiably}. Action projection methods project the unsafe actions to close safe actions~\cite{kochdumper2022provably}.

Control theory based SRRL approaches have utilized Lyapunov functions and Model Predictive Control (MPC). Chow \textit{et al.}~\cite{chow2018lyapunov, chow2019lyapunov} introduced  Lyapunov functions based on discrete and continuous control methods for the global safety of behavior policy in RL. However, designing Lyapunov functions for different environments may be difficult, and the requirement of a baseline policy may be challenging to satisfy during real-world applications. Koller \textit{et al.}~\cite{koller2018learning} proposed an MPC based method to ensure safe exploration using a statistical model of the system. Marvi and  Kiumarsi~\cite{marvi2021safe}  introduced a control barrier function method for safe off-policy robot learning without needing to have a thorough understanding of system dynamics, in which the cost functions are augmented by a control barrier function.


Most of the above methods do not consider human-robot interaction. There are some prior works that investigate human-centered SRRL. For instance,  Kazantzidis1 \textit{et al.}~\cite{kazantzidis2022train} introduced a mechanism to ensure safety during exploration by harnessing human preferences. Reddy \textit{et al.}~\cite{reddy2020learning} present a method to learn the model of human objectives by leveraging human feedback based on hypothetical behaviors, and then the model can be used to ensure the safety of robot learning. Saunders \textit{et al.}~\cite{saunders2017trial} try to guarantee reinforcement learning safety by human interventions, where human interventions are learned through a supervised learning model. However, most of these works do not consider the mutual influence of humans and robots in shared environments. Thus, interactive behaviors between humans and robots that leverage bi-directional information transfer, as shown in Figure~\ref{fig:safe-robot-rl-interactive-bahvaiors}, still need further investigation to ensure SRRL.

\section{Human-centered Safe Robot Reinforcement Learning Framework}
 \begin{figure}[htbp!]
  \centering
  \includegraphics[width=0.95\linewidth]{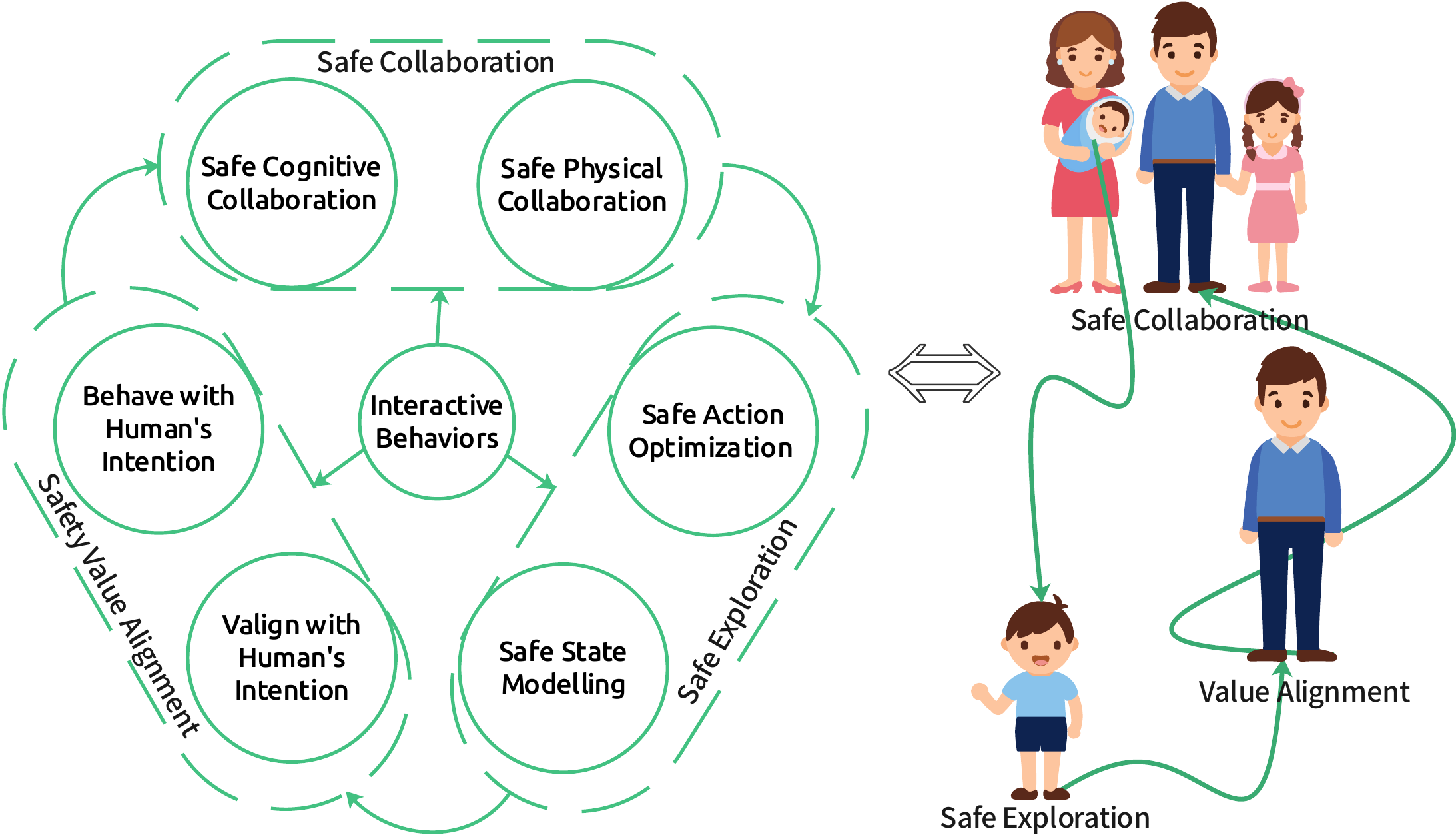}
  \caption{Three stages of human-centered SRRL. When a newborn infant enters a new environment, if it wants to survive in the new environment, it must first learn about the environment via interactive behaviors (\textbf{Safe Exploration}), e.g., bi-directional feedback with its parents to learn how to walk. Then it should find something helpful to survive better via interactive behaviors (\textbf{Value Alignment}), e.g., bi-directional feedback with its teachers to learn human values. Finally, it has to collaborate with others to maximize the society's reward via interactive behaviors (\textbf{Safe Collaboration}), e.g., bi-directional feedback with its collaborators to create new things.}
  \label{fig:framework-safe-robot-rl}
\end{figure}

Our proposed human-centered SRRL framework, as shown in Figure~\ref{fig:framework-safe-robot-rl}, consists of three stages: safe exploration, safety value alignment, and safe collaboration. When a robot enters a new environment, it must explore it through safe action optimization and state modeling. Then, the robot has to learn the safety value alignment from the human-robot interaction. Finally, the robot should be able to achieve safe collaboration with humans. Interactive behaviors can help successfully achieve each of these three stages.
 


\subsection{Safe Exploration}





A robot should explore and find helpful information when entering a new environment. During the exploration, the robot must ensure its and the environment's (including other agents') safety. Therefore, the first stage of human-centered SRRL, safe exploration, focuses on exploring new environments and getting information about the environment model through safe actions. Uncertainty about the environment makes it difficult to determine safe actions. For example, newborn infants sometimes end up putting harmful objects in their mouths during oral exploration. It is challenging to ensure safe actions using conventional RL methods that learn new skills through trial and error. Especially with model-free RL, a robot cannot avoid a destructive action unless it has already tried it (or a similar action)~\cite{saunders2017trial}.



Several methods have been proposed to address the challenge of safe exploration for RL~\cite{bharadhwaj2021conservative, garcia2012safe, liu2022robot, liu2020robust, turchetta2021safety, xiong2021safety, liu2022robustness}. For example, Garcia and Fern{\'a}ndez~\cite{garcia2012safe} introduced a safe exploration method, in which a predefined baseline policy is required, and the baseline policy is approximated by behavioral cloning methods~\cite{anderson2000behavioral}. Nonetheless, derived from the method, it would be hard to search for the optimal exploration policy, and the capabilities of the baseline policy could severely limit the method's performance. A conservative safety critic \cite{bharadhwaj2021conservative} is proposed to guarantee safety with high probability during robot exploration processes. Garcia and Fern{\'a}ndez \cite{garcia2012safe} introduced a smoother risk function for safe robot exploration, which can achieve monotonic reward improvement and ensure safety. This method needs a predefined baseline policy to explore the safe space. Liu \textit{ et al.} \cite{liu2022robot} developed a safe exploration method for robot learning by constructing a constrained manifold. This method can guarantee safety for robot exploration using model-free RL. However, it requires an accurate robot model or a perfect tracking controller, which may hinder their method's real-world applications. 

Interactive behaviors can be used to transfer expert knowledge from the human to the robot for safe exploration, just like infants are safeguarded by their parents. Human intervention has been investigated to avoid catastrophes in RL~\cite{saunders2017trial}, and human feedback can be a reference for RL to ensure robot safety during safe exploration~\cite{frye2019parenting}. However, the amount of human labor required for complex real-world applications is prohibitive. The bi-directional feedback in interactive behaviors can reduce the human-time cost during safe exploration. On the one hand, the robot can actively query the human and provide explanations of its behavior. On the other hand, the human can also actively query the robot to learn about the robot's model, in addition to intervening for safe exploration.

\subsection{Safety Value Alignment}
In the second stage of safe robot learning, to train a robot to perform a task safely, we need to evaluate how well it performs in terms of its performance on safety. Whether it is a form of costs, rewards, or labels, we need some form of signals to guide the safety policies for robot learning. 
In some scenarios, the safety performance can be evaluated automatically, such as bumping into other vehicles of autonomous driving or breaking the arms in robot manipulation. In these cases, training signals can be straightforwardly defined for safe robot learning. Furthermore, to facilitate the safe deployment of agents in real-world tasks, agents also need to be compatible with users' ethical judgment. Taking automated vehicles as an example, should one autonomous vehicle cut in line to maximize its reward in achieving a goal? Instead of maximizing only the reward, the agents need the capacity to abide by human moral values, which is essential but lacks effort.

Current robot learning algorithms rely on humans to state these training signals \cite{gu2022review, yuan2022situ} and assume that humans understand the dangers. For example, imitation learning inferences a reward function from human demonstrations; preference-based learning guides the robot based on human judgments. These classes of tasks involve “human” training signals. 
The related problem of how to align has been discussed in earlier literature \cite{leike2018scalable,christiano2018supervising,liu2022MRN,kazantzidis2022train} on how to align agents with user intentions, in which meaningful training signals can be hard to obtain, due to the unpredictable long-term effect of the behaviors, or potential influence to other agents and environments in large multi-agent systems. 



{Designing AI agents that can achieve arbitrary objectives, such as minimizing some cost or penalties, can be deficient in that the systems are intrinsically unpredictable and might result in negative and irreversible outcomes for humans. In the context of interactive learning, we consider how a robot can behave safely or align with the user's intentions whilst maintaining safety under interactive behaviors with humans, and we frame this as the \textbf{safety value alignment} problem: \textit{ how to create robots that behave safely and  align  with the human's intentions?}}

{ 

Interactive behaviors allow agents to infer human values. While agents infer human values from their feedback, bi-directional feedback enables the agent to explain its decision-making process. 
One early attempt \cite{yuan2022situ} studied bi-directional communication in tabular-based navigation tasks without considering more practical scenarios. The next step aims to study the generalization of such results to large-scale problems via more efficient algorithms.

}



\subsection{Safe Human-Robot Collaboration}
The third stage of our safe-robot learning framework aims to accomplish safe physical and cognitive collaboration between robots and humans. Collaboration has enabled humans to achieve great evolutionary success. Therefore, safe human-robot collaboration is essential for successful human-robot co-existence.  

The four main categories of human-robot collaboration tasks explored in the literature are collaborative assembly, object handling, object handovers, and collaborative manufacturing~\cite{semeraro2023human}. RL has been used for tuning impedance controllers in physical human-robot collaboration tasks such as lifting objects~\cite{roveda2020model} and guided trajectory following~\cite{modares2015optimized}. However, these works evaluated the controllers in simplified scenarios and did not evaluate the generalizability of the learned policies. RL has also been used for performing robot-to-human object handovers~\cite{kupcsik2018learning} and human-to-robot object handovers~\cite{chang2022learning}. Nevertheless, in some cases, the spatial generalizability of learned policies is low~\cite{kshirsagar2021evaluating}. Ghadirzadeh et al.~\cite{ghadirzadeh2020human} used deep q-learning to generate proactive robot actions in a human-robot collaborative packaging task. However, they only evaluated a specific task scenario with a highly engineered reward function. Also, they did not test the trained policy in the real world and for different human participants than the training set. 

Deep RL methods have been applied in real-world learning scenarios for tasks like quadrupedal walking, grasping objects, and varied manipulation skills~\cite{ibarz2021train}. One of the desired features of these works is the ability to perform training with little or no human involvement. However, scenarios of human-robot collaboration typically involve multiple humans in the robot's learning process. Multi-agent RL methods such as self-play or population-play do not perform very well with human partners~\cite{carroll2019utility}. One proposed solution called Fictitious Co-Play (FCP) involves training with a population of self-play agents and their past checkpoints taken throughout training~\cite{strouse2021collaborating}. However, FCP was evaluated only in a virtual game environment.

Interactive RL (IRL) approaches involve a human-in-the-loop to guide the robot's RL process. IRL has been applied to various human-computer interaction scenarios~\cite{arzate2020survey}. In addition, human social feedback in the form of evaluation, advice, or instruction has also been utilized for several robot RL tasks~\cite{lin2020review}. However, more research is needed towards utilizing IRL for safe human-robot collaboration. Also, while some works have explored non-verbal cues to express the robot's uncertainty during the learning process~\cite{matarese2021toward}, most existing IRL approaches do not involve feedback from robots to humans. As depicted in Figure~\ref{fig:framework-safe-robot-rl}, evaluative feedback from robots to humans could help improve human-robot collaboration.














\section{Open Challenges}

In this section, we describe four key open challenges towards utilizing interactive behaviors for SRRL. These open challenges are related to the robustness, efficiency, transparency, and adaptability of SRRL.
\begin{enumerate}
    \item How can the robot learn robust behaviours with potential human adversaries?
    \item How to improve data efficiency of SRRL for effective utilization of interactive behaviors?
    \item How to design ``transparent" user interfaces for interactive behaviors?
    \item How to enhance adaptability of SRRL for handling multiple scenarios of interactive behaviors?
    
\end{enumerate}

The first challenge is to achieve robust SRRL with respect to unintentional or intentionally erroneous human conduct in interactive behaviors. In existing SRRL methods, it is often neglected that humans might misstate the safety signals. Also, due to the potential involvement of multiple humans with different values, robots need to learn to strike a balance between them. In the extreme case, adversaries may intentionally state their signal to mislead the training of robots to achieve malicious objectives. Training robust agents against such malicious users needs further research. Adversarial training may be useful to ensure safety in such scenarios~\cite{meng2022integrating}. However, adversarial training is not yet ready for real-world robot learning~\cite{lechner2021adversarial}.  




The second challenge is to improve the data efficiency of SRRL, given that real-world interactive behaviors are expensive. Data efficiency can determine how quickly robots learn new skills and adapt to new environments and how effectively interactive behaviors can be utilized in the learning process. The success of machine learning can be attributed to the availability of large datasets and simulation environments. Therefore, one possible solution to reduce the need for real-world interactive data is to build large datasets or simulations of interactive behaviors. For example, Lee et al.\cite{lee2022towards} present a mixed reality (MR) framework in which humans can interact with virtual robots in virtual or augmented reality (VR/AR) environments. It can serve as a platform for collecting data in various human-robot interaction and collaboration scenarios. However, this framework suffers from the limiting aspects of MR, such as the inconvenience of wearable interfaces, motion sickness, and fatigue. Improvements in MR technology will be crucial for the widespread use of such MR environments. 




The third challenge is to maintain transparency during SRRL. Transparency is important for the effective utilization of interactive behaviors. James \textit{et al.} \cite{macglashan2017interactive} provide empirical results to demonstrate human feedback and robot policy can be interactively influenced by each other, and indicate that the assumption is that the feedback from a human is independent of the robot's current policy, may be incorrect. Transparency can be achieved in the context of interactive behaviors by explaining the robot's decision-making process to the human and exposing the robot's internal state. Several solutions have been proposed to improve the explainability of robot RL~\cite{macglashan2017interactive, matarese2021toward, hayes2017improving, likmeta2020combining, van2018contrastive, atakishiyev2021towards, atakishiyev2021explainable}. 
For example, Likmeta \textit{et al.}~\cite{likmeta2020combining} introduced an interpretable rule-based controller for the transparency of RL in transportation applications. Nonetheless, the policy that the method provides may be too conservative, since it severely depends on the restricted rules. Matarese \textit{et al.}~\cite{matarese2021toward} present a method to improve robot behaviors' transparency to human users by leveraging emotional-behavior feedback based on robot learning progress. However, further research is needed towards communicating the robot's internal state. Non-verbal communication in the form of gazes and gestures can be leveraged to communicate the robot's internal state. For example, if the robot is uncertain about its decisions, it can show hesitation gestures. 


\begin{figure}[htbp!]
  \centering
  \includegraphics[width=0.35\linewidth]{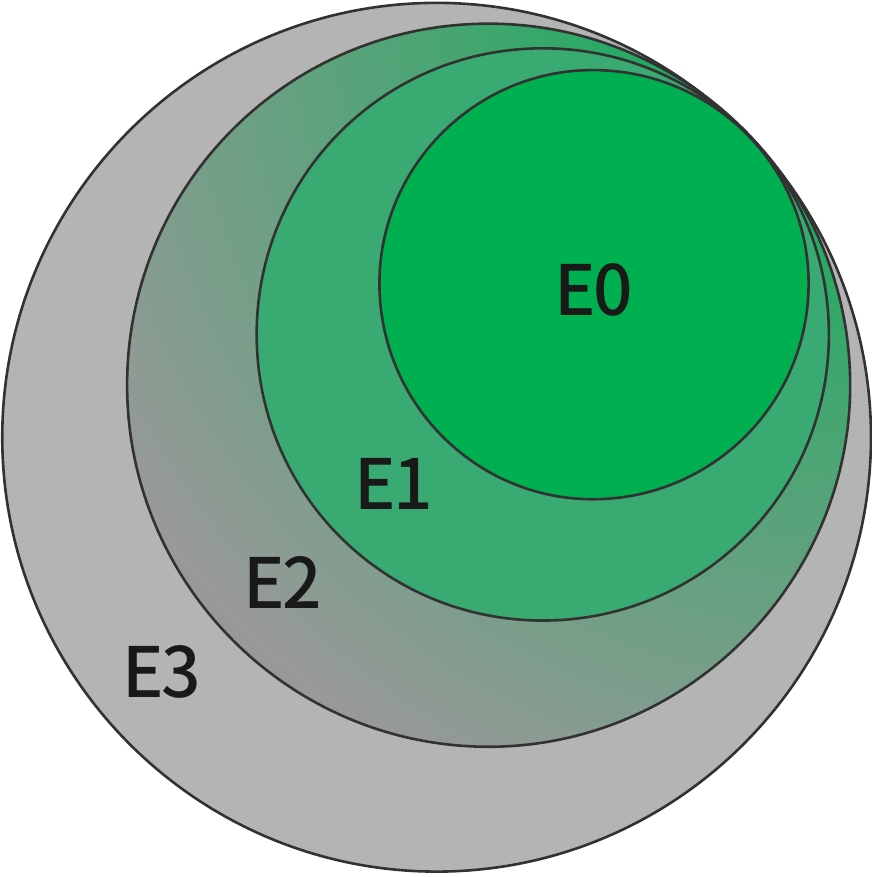}
  \caption{Trusted Environments (E0) and "3U" Environments (E1: Unstable, E2: Uncertain, E3: Unknown). As the color gradually changes from gray to green, the information observed is more and more abundant, and the environment is more and more certain.}
  \label{fig:environments-definitions-safe-robot}
\end{figure}

The fourth challenge is to enhance the adaptability of safe robot learning to handle a variety of settings involving interactive behaviors. As shown in Figure \ref{fig:environments-definitions-safe-robot}, the robot can encounter different environments. $E0$ denotes completely trusted environments, in which the environment information is known and the environment is stable; $E1$ denotes \textbf{U}nstable environments for which environment information is known; $E2$ represents \textbf{U}ncertain environments, where the information about the environment is uncertain and partial; $E3$ represents \textbf{U}nknown environments, in which the environment information is completely unknown.  

Safety can be ensured in trusted environments, e.g., robots can safely grasp an object in a static environment. However, in ``3U" environments, ensuring SRRL is challenging. For instance, during sim-to-real transfer,  the discrepancies between the simulation models and real-world models are inevitable~\cite{mitsch2016modelplex}, and the real-world environments are replete with uncertain disturbances and unknown information, e.g., in a multi-agent system, guaranteeing each agent's safety may be difficult. Some works provide a potential direction to ensure multi-robot learning safety in unstable environments, for example, Multi-Agent Constrained Policy Optimization and Multi-Agent Proximal Policy Optimization Lagrangian~\cite{gu2021multi}. Nevertheless, the exploration involving interactive behaviors between agents and environments can be intricate and time-intensive. The incorporation of human insights and value alignment in the exploratory phase is instrumental in enhancing the adaptability of these agents within a human-in-the-loop learning system. As we envisage future trajectories of research, a salient focus is the attainment of SRRL characterized by interactive behaviors in “3U” environments. In this vein, game theory ~\cite{fudenberg1991game} emerges as a pivotal tool, offering nuanced strategies and frameworks for optimizing agent-environment interactions. Concurrently, the integration of advancements in cognitive science is anticipated to play a quintessential role. Specific areas of interest encompass the optimization of information management protocols between humans and robotic agents and the augmentation of robotic cognitive faculties through the effective utilization of perception devices. These integrated approaches aim to engender a more seamless, efficient, and adaptive interaction paradigm, catalyzing enhanced performance and adaptability in complex, dynamic environments.

However, the exploration with interactive behaviors between agents and environments may be time-consuming. Involving human knowledge and value alignment in the exploration can improve its adaptability in a human-loop learning system. Future work to achieve SRRL with interactive behaviors in ``3U" environments can leverage game theory~\cite{fudenberg1991game} and advances in cognitive science, for instance, how to manage the information between humans and robots, and robots how to leverage perception devices to enhance its cognitive abilities.

\section{Conclusion}
  Robots need to ensure safety while leveraging RL in human environments. However, conventional SRRL algorithms do not consider the mutual influence between the robot and the human. In this paper, we proposed a human-centered SRRL framework consisting of three stages: safe exploration, value alignment, and human-robot collaboration. We discussed how these stages can leverage mutual influence or bidirectional information transfer between the robot and the human through interactive behaviors. Furthermore, we described four key open challenges related to the robustness, efficiency, transparency, and adaptability of SRRL for effective utilization of interactive behaviors. 


 \section*{Acknowledgments}
 We would like to
thank Yaodong Yang for his useful suggestions.
 


\normalem






\bibliography{main}{}

\begin{thebibliography}{}

\end{thebibliography}


\begin{thebibliography}{10}

\bibitem{achiam2017constrained}
Joshua Achiam, David Held, Aviv Tamar, and Pieter Abbeel.
\newblock Constrained policy optimization.
\newblock In {\em International conference on machine learning}, pages 22--31. PMLR, 2017.

\bibitem{akametalu2014reachability}
Anayo~K Akametalu, Jaime~F Fisac, Jeremy~H Gillula, Shahab Kaynama, Melanie~N Zeilinger, and Claire~J Tomlin.
\newblock Reachability-based safe learning with gaussian processes.
\newblock In {\em 53rd IEEE Conference on Decision and Control}, pages 1424--1431. IEEE, 2014.

\bibitem{anderson2000behavioral}
Charles~W Anderson, Bruce~A Draper, and David~A Peterson.
\newblock Behavioral cloning of student pilots with modular neural networks.
\newblock In {\em ICML}, pages 25--32, 2000.

\bibitem{arzate2020survey}
Christian Arzate~Cruz and Takeo Igarashi.
\newblock A survey on interactive reinforcement learning: Design principles and open challenges.
\newblock In {\em Proceedings of the 2020 ACM designing interactive systems conference}, pages 1195--1209, 2020.

\bibitem{atakishiyev2021explainable}
Shahin Atakishiyev, Mohammad Salameh, Hengshuai Yao, and Randy Goebel.
\newblock Explainable artificial intelligence for autonomous driving: A comprehensive overview and field guide for future research directions.
\newblock {\em arXiv preprint arXiv:2112.11561}, 2021.

\bibitem{atakishiyev2021towards}
Shahin Atakishiyev, Mohammad Salameh, Hengshuai Yao, and Randy Goebel.
\newblock Towards safe, explainable, and regulated autonomous driving.
\newblock {\em arXiv preprint arXiv:2111.10518}, 2021.

\bibitem{baumann2021gosafe}
Dominik Baumann, Alonso Marco, Matteo Turchetta, and Sebastian Trimpe.
\newblock Gosafe: Globally optimal safe robot learning.
\newblock In {\em 2021 IEEE International Conference on Robotics and Automation (ICRA)}, pages 4452--4458. IEEE, 2021.

\bibitem{berkenkamp2016safe}
Felix Berkenkamp, Riccardo Moriconi, Angela~P Schoellig, and Andreas Krause.
\newblock Safe learning of regions of attraction for uncertain, nonlinear systems with gaussian processes.
\newblock In {\em 2016 IEEE 55th Conference on Decision and Control (CDC)}, pages 4661--4666. IEEE, 2016.

\bibitem{bharadhwaj2021conservative}
Homanga Bharadhwaj, Aviral Kumar, Nicholas Rhinehart, Sergey Levine, Florian Shkurti, and Animesh Garg.
\newblock Conservative safety critics for exploration.
\newblock In {\em International Conference on Learning Representations (ICLR)}, 2021.

\bibitem{brunke2021safe}
Lukas Brunke, Melissa Greeff, Adam~W Hall, Zhaocong Yuan, Siqi Zhou, Jacopo Panerati, and Angela~P Schoellig.
\newblock Safe learning in robotics: From learning-based control to safe reinforcement learning.
\newblock {\em Annual Review of Control, Robotics, and Autonomous Systems}, 5, 2021.

\bibitem{carroll2019utility}
Micah Carroll, Rohin Shah, Mark~K Ho, Tom Griffiths, Sanjit Seshia, Pieter Abbeel, and Anca Dragan.
\newblock On the utility of learning about humans for human-ai coordination.
\newblock {\em Advances in neural information processing systems}, 32, 2019.

\bibitem{chang2022learning}
Po-Kai Chang, Jui-Te Huang, Yu-Yen Huang, and Hsueh-Cheng Wang.
\newblock Learning end-to-end 6dof grasp choice of human-to-robot handover using affordance prediction and deep reinforcement learning.
\newblock In {\em 2022 IEEE International Conference on Robotics and Automation (ICRA). IEEE}, 2022.

\bibitem{chow2018lyapunov}
Yinlam Chow, Ofir Nachum, Edgar Duenez-Guzman, and Mohammad Ghavamzadeh.
\newblock A lyapunov-based approach to safe reinforcement learning.
\newblock {\em Advances in neural information processing systems}, 31, 2018.

\bibitem{chow2019lyapunov}
Yinlam Chow, Ofir Nachum, Aleksandra Faust, Edgar Duenez-Guzman, and Mohammad Ghavamzadeh.
\newblock Lyapunov-based safe policy optimization for continuous control.
\newblock {\em arXiv preprint arXiv:1901.10031}, 2019.

\bibitem{christiano2018supervising}
Paul Christiano, Buck Shlegeris, and Dario Amodei.
\newblock Supervising strong learners by amplifying weak experts.
\newblock {\em arXiv preprint arXiv:1810.08575}, 2018.

\bibitem{du2019liir}
Yali Du, Lei Han, Meng Fang, Tianhong Dai, Ji~Liu, and Dacheng Tao.
\newblock Liir: learning individual intrinsic reward in multi-agent reinforcement learning.
\newblock In {\em Proceedings of the 33rd International Conference on Neural Information Processing Systems (NeurIPS)}, pages 4403--4414, 2019.

\bibitem{frye2019parenting}
Christopher Frye and Ilya Feige.
\newblock Parenting: Safe reinforcement learning from human input.
\newblock {\em arXiv preprint arXiv:1902.06766}, 2019.

\bibitem{fudenberg1991game}
Drew Fudenberg and Jean Tirole.
\newblock {\em Game theory}.
\newblock MIT press, 1991.

\bibitem{fulton2018safe}
Nathan Fulton and Andr{\'e} Platzer.
\newblock Safe reinforcement learning via formal methods: Toward safe control through proof and learning.
\newblock In {\em Proceedings of the AAAI Conference on Artificial Intelligence}, volume~32, 2018.

\bibitem{gao2022scaling}
Leo Gao, John Schulman, and Jacob Hilton.
\newblock Scaling laws for reward model overoptimization.
\newblock {\em arXiv preprint arXiv:2210.10760}, 2022.

\bibitem{garcia2012safe}
Javier Garcia and Fernando Fern{\'a}ndez.
\newblock Safe exploration of state and action spaces in reinforcement learning.
\newblock {\em Journal of Artificial Intelligence Research}, 45:515--564, 2012.

\bibitem{ghadirzadeh2020human}
Ali Ghadirzadeh, Xi~Chen, Wenjie Yin, Zhengrong Yi, M{\aa}rten Bj{\"o}rkman, and Danica Kragic.
\newblock Human-centered collaborative robots with deep reinforcement learning.
\newblock {\em IEEE Robotics and Automation Letters}, 6(2):566--571, 2020.

\bibitem{gu2022constrained}
Shangding Gu, Guang Chen, Lijun Zhang, Jing Hou, Yingbai Hu, and Alois Knoll.
\newblock Constrained reinforcement learning for vehicle motion planning with topological reachability analysis.
\newblock {\em Robotics}, 11(4):81, 2022.

\bibitem{gu2021multi}
Shangding Gu, Jakub~Grudzien Kuba, Yuanpei Chen, Yali Du, Long Yang, Alois Knoll, and Yaodong Yang.
\newblock Safe multi-agent reinforcement learning for multi-robot control.
\newblock {\em Artificial Intelligence}, 319:103905, 2023.

\bibitem{gu2022review}
Shangding Gu, Long Yang, Yali Du, Guang Chen, Florian Walter, Jun Wang, Yaodong Yang, and Alois Knoll.
\newblock A review of safe reinforcement learning: Methods, theory and applications.
\newblock {\em arXiv preprint arXiv:2205.10330}, 2022.

\bibitem{han2019grid}
Lei Han, Peng Sun, Yali Du, Jiechao Xiong, Qing Wang, Xinghai Sun, Han Liu, and Tong Zhang.
\newblock Grid-wise control for multi-agent reinforcement learning in video game ai.
\newblock In {\em International Conference on Machine Learning (ICML)}, pages 2576--2585. PMLR, 2019.

\bibitem{hayes2017improving}
Bradley Hayes and Julie~A Shah.
\newblock Improving robot controller transparency through autonomous policy explanation.
\newblock In {\em 2017 12th ACM/IEEE International Conference on Human-Robot Interaction (HRI}, pages 303--312. IEEE, 2017.

\bibitem{hunt2021verifiably}
Nathan Hunt, Nathan Fulton, Sara Magliacane, Trong~Nghia Hoang, Subhro Das, and Armando Solar-Lezama.
\newblock Verifiably safe exploration for end-to-end reinforcement learning.
\newblock In {\em Proceedings of the 24th International Conference on Hybrid Systems: Computation and Control}, pages 1--11, 2021.

\bibitem{ibarz2021train}
Julian Ibarz, Jie Tan, Chelsea Finn, Mrinal Kalakrishnan, Peter Pastor, and Sergey Levine.
\newblock How to train your robot with deep reinforcement learning: lessons we have learned.
\newblock {\em The International Journal of Robotics Research}, 40(4-5):698--721, 2021.

\bibitem{kaushik2022safeapt}
Rituraj Kaushik, Karol Arndt, and Ville Kyrki.
\newblock Safeapt: Safe simulation-to-real robot learning using diverse policies learned in simulation.
\newblock {\em IEEE Robotics and Automation Letters}, 2022.

\bibitem{kazantzidis2022train}
Ilias Kazantzidis, Timothy~J Norman, Yali Du, and Christopher~T Freeman.
\newblock How to train your agent: Active learning from human preferences and justifications in safety-critical environments.
\newblock In {\em Proceedings of the 21st International Conference on Autonomous Agents and Multiagent Systems}, pages 1654--1656, 2022.

\bibitem{knox2009interactively}
W~Bradley Knox and Peter Stone.
\newblock Interactively shaping agents via human reinforcement: The tamer framework.
\newblock In {\em Proceedings of the fifth international conference on Knowledge capture}, pages 9--16, 2009.

\bibitem{kochdumper2022provably}
Niklas Kochdumper, Hanna Krasowski, Xiao Wang, Stanley Bak, and Matthias Althoff.
\newblock Provably safe reinforcement learning via action projection using reachability analysis and polynomial zonotopes.
\newblock {\em arXiv preprint arXiv:2210.10691}, 2022.

\bibitem{koller2018learning}
Torsten Koller, Felix Berkenkamp, Matteo Turchetta, and Andreas Krause.
\newblock Learning-based model predictive control for safe exploration.
\newblock In {\em 2018 IEEE conference on decision and control (CDC)}, pages 6059--6066. IEEE, 2018.

\bibitem{kong2018human}
Xiangjie Kong, Kai Ma, Shen Hou, Di~Shang, and Feng Xia.
\newblock Human interactive behavior: A bibliographic review.
\newblock {\em IEEE Access}, 7:4611--4628, 2018.

\bibitem{krasowski2022provably}
Hanna Krasowski, Jakob Thumm, Marlon M{\"u}ller, Xiao Wang, and Matthias Althoff.
\newblock Provably safe reinforcement learning: A theoretical and experimental comparison.
\newblock {\em arXiv preprint arXiv:2205.06750}, 2022.

\bibitem{krasowski2020safe}
Hanna Krasowski, Xiao Wang, and Matthias Althoff.
\newblock Safe reinforcement learning for autonomous lane changing using set-based prediction.
\newblock In {\em 2020 IEEE 23rd International Conference on Intelligent Transportation Systems (ITSC)}, pages 1--7. IEEE, 2020.

\bibitem{kroemer2021review}
Oliver Kroemer, Scott Niekum, and George Konidaris.
\newblock A review of robot learning for manipulation: Challenges, representations, and algorithms.
\newblock {\em The Journal of Machine Learning Research}, 22(1):1395--1476, 2021.

\bibitem{kshirsagar2021evaluating}
Alap Kshirsagar, Guy Hoffman, and Armin Biess.
\newblock Evaluating guided policy search for human-robot handovers.
\newblock {\em IEEE Robotics and Automation Letters}, 6(2):3933--3940, 2021.

\bibitem{kupcsik2018learning}
Andras Kupcsik, David Hsu, and Wee~Sun Lee.
\newblock Learning dynamic robot-to-human object handover from human feedback.
\newblock In {\em Robotics research}, pages 161--176. Springer, 2018.

\bibitem{lechner2021adversarial}
Mathias Lechner, Ramin Hasani, Radu Grosu, Daniela Rus, and Thomas~A Henzinger.
\newblock Adversarial training is not ready for robot learning.
\newblock In {\em 2021 IEEE International Conference on Robotics and Automation (ICRA)}, pages 4140--4147. IEEE, 2021.

\bibitem{lecun2015deep}
Yann LeCun, Yoshua Bengio, and Geoffrey Hinton.
\newblock Deep learning.
\newblock {\em nature}, 521(7553):436--444, 2015.

\bibitem{lee2022towards}
Bhoram Lee, Jonathan Brookshire, Rhys Yahata, and Supun Samarasekera.
\newblock Towards safe, realistic testbed for robotic systems with human interaction.
\newblock In {\em 2022 International Conference on Robotics and Automation (ICRA)}, pages 11280--11287. IEEE, 2022.

\bibitem{leike2018scalable}
Jan Leike, David Krueger, Tom Everitt, Miljan Martic, Vishal Maini, and Shane Legg.
\newblock Scalable agent alignment via reward modeling: a research direction.
\newblock {\em arXiv preprint arXiv:1811.07871}, 2018.

\bibitem{li2019temporal}
Xiao Li and Calin Belta.
\newblock Temporal logic guided safe reinforcement learning using control barrier functions.
\newblock {\em arXiv preprint arXiv:1903.09885}, 2019.

\bibitem{likmeta2020combining}
Amarildo Likmeta, Alberto~Maria Metelli, Andrea Tirinzoni, Riccardo Giol, Marcello Restelli, and Danilo Romano.
\newblock Combining reinforcement learning with rule-based controllers for transparent and general decision-making in autonomous driving.
\newblock {\em Robotics and Autonomous Systems}, 131:103568, 2020.

\bibitem{lin2020review}
Jinying Lin, Zhen Ma, Randy Gomez, Keisuke Nakamura, Bo~He, and Guangliang Li.
\newblock A review on interactive reinforcement learning from human social feedback.
\newblock {\em IEEE Access}, 8:120757--120765, 2020.

\bibitem{liu2020robust}
Anqi Liu, Guanya Shi, Soon-Jo Chung, Anima Anandkumar, and Yisong Yue.
\newblock Robust regression for safe exploration in control.
\newblock In {\em Learning for Dynamics and Control}, pages 608--619. PMLR, 2020.

\bibitem{liu2022robot}
Puze Liu, Davide Tateo, Haitham~Bou Ammar, and Jan Peters.
\newblock Robot reinforcement learning on the constraint manifold.
\newblock In {\em Conference on Robot Learning}, pages 1357--1366. PMLR, 2022.

\bibitem{liu2022MRN}
Runze Liu, Fengshuo Bai, Yali Du, and Yaodong Yang.
\newblock Meta-reward-net: Implicitly differentiable reward learning for preference-based reinforcement learning.
\newblock In {\em Advances in Neural Information Processing Systems (NeurIPS)}, 2022.

\bibitem{liu2022robustness}
Zuxin Liu, Zijian Guo, Zhepeng Cen, Huan Zhang, Jie Tan, Bo~Li, and Ding Zhao.
\newblock On the robustness of safe reinforcement learning under observational perturbations.
\newblock {\em arXiv preprint arXiv:2205.14691}, 2022.

\bibitem{lou2023pecan}
Xingzhou Lou, Jiaxian Gu, Junge Zhang, Jun Wang, Kaiqi Huang, and Yali Du.
\newblock Pecan: Leveraging policy ensemble for context-aware zero-shot human-ai coordination.
\newblock In {\em Proceedings of the 22st International Conference on Autonomous Agents and Multiagent Systems (AAMAS)}, pages 1654--1666, 2023.

\bibitem{macglashan2017interactive}
James MacGlashan, Mark~K Ho, Robert Loftin, Bei Peng, Guan Wang, David~L Roberts, Matthew~E Taylor, and Michael~L Littman.
\newblock Interactive learning from policy-dependent human feedback.
\newblock In {\em International Conference on Machine Learning}, pages 2285--2294. PMLR, 2017.

\bibitem{marco2021robot}
Alonso Marco, Dominik Baumann, Majid Khadiv, Philipp Hennig, Ludovic Righetti, and Sebastian Trimpe.
\newblock Robot learning with crash constraints.
\newblock {\em IEEE Robotics and Automation Letters}, 6(2):1439--1446, 2021.

\bibitem{marvi2021safe}
Zahra Marvi and Bahare Kiumarsi.
\newblock Safe reinforcement learning: A control barrier function optimization approach.
\newblock {\em International Journal of Robust and Nonlinear Control}, 31(6):1923--1940, 2021.

\bibitem{matarese2021toward}
Marco Matarese, Alessandra Sciutti, Francesco Rea, and Silvia Rossi.
\newblock Toward robots’ behavioral transparency of temporal difference reinforcement learning with a human teacher.
\newblock {\em IEEE Transactions on Human-Machine Systems}, 51(6):578--589, 2021.

\bibitem{meng2022integrating}
Jinling Meng, Fei Zhu, Yangyang Ge, and Peiyao Zhao.
\newblock Integrating safety constraints into adversarial training for robust deep reinforcement learning.
\newblock {\em Information Sciences}, 2022.

\bibitem{mitsch2016modelplex}
Stefan Mitsch and Andr{\'e} Platzer.
\newblock Modelplex: Verified runtime validation of verified cyber-physical system models.
\newblock {\em Formal Methods in System Design}, 49(1):33--74, 2016.

\bibitem{modares2015optimized}
Hamidreza Modares, Isura Ranatunga, Frank~L Lewis, and Dan~O Popa.
\newblock Optimized assistive human--robot interaction using reinforcement learning.
\newblock {\em IEEE transactions on cybernetics}, 46(3):655--667, 2015.

\bibitem{OpenAI_gpt}
OpenAI.
\newblock Chatgpt.
\newblock https://openai.com/blog/chatgpt/, 2023.

\bibitem{reddy2020learning}
Siddharth Reddy, Anca Dragan, Sergey Levine, Shane Legg, and Jan Leike.
\newblock Learning human objectives by evaluating hypothetical behavior.
\newblock In {\em International Conference on Machine Learning}, pages 8020--8029. PMLR, 2020.

\bibitem{roveda2020model}
Loris Roveda, Jeyhoon Maskani, Paolo Franceschi, Arash Abdi, Francesco Braghin, Lorenzo Molinari~Tosatti, and Nicola Pedrocchi.
\newblock Model-based reinforcement learning variable impedance control for human-robot collaboration.
\newblock {\em Journal of Intelligent \& Robotic Systems}, 100(2):417--433, 2020.

\bibitem{saunders2017trial}
William Saunders, Girish Sastry, Andreas Stuhlm{\"u}ller, and Owain Evans.
\newblock Trial without error: Towards safe reinforcement learning via human intervention.
\newblock In {\em Proceedings of the 17th International Conference on Autonomous Agents and MultiAgent Systems}, pages 2067--2069, 2018.

\bibitem{semeraro2023human}
Francesco Semeraro, Alexander Griffiths, and Angelo Cangelosi.
\newblock Human--robot collaboration and machine learning: A systematic review of recent research.
\newblock {\em Robotics and Computer-Integrated Manufacturing}, 79:102432, 2023.

\bibitem{silver2018general}
David Silver, Thomas Hubert, Julian Schrittwieser, Ioannis Antonoglou, Matthew Lai, Arthur Guez, Marc Lanctot, Laurent Sifre, Dharshan Kumaran, Thore Graepel, et~al.
\newblock A general reinforcement learning algorithm that masters chess, shogi, and go through self-play.
\newblock {\em Science}, 362(6419):1140--1144, 2018.

\bibitem{stiennon2020learning}
Nisan Stiennon, Long Ouyang, Jeffrey Wu, Daniel Ziegler, Ryan Lowe, Chelsea Voss, Alec Radford, Dario Amodei, and Paul~F Christiano.
\newblock Learning to summarize with human feedback.
\newblock {\em Advances in Neural Information Processing Systems}, 33:3008--3021, 2020.

\bibitem{strouse2021collaborating}
DJ~Strouse, Kevin McKee, Matt Botvinick, Edward Hughes, and Richard Everett.
\newblock Collaborating with humans without human data.
\newblock {\em Advances in Neural Information Processing Systems}, 34:14502--14515, 2021.

\bibitem{sui2015safe}
Yanan Sui, Alkis Gotovos, Joel Burdick, and Andreas Krause.
\newblock Safe exploration for optimization with gaussian processes.
\newblock In {\em International conference on machine learning}, pages 997--1005. PMLR, 2015.

\bibitem{sui2018stagewise}
Yanan Sui, Vincent Zhuang, Joel Burdick, and Yisong Yue.
\newblock Stagewise safe bayesian optimization with gaussian processes.
\newblock In {\em International conference on machine learning}, pages 4781--4789. PMLR, 2018.

\bibitem{sutton2018reinforcement}
Richard~S Sutton and Andrew~G Barto.
\newblock {\em Reinforcement learning: An introduction}.
\newblock MIT press, 2018.

\bibitem{tamar2012policy}
Aviv Tamar, Dotan Di~Castro, and Shie Mannor.
\newblock Policy gradients with variance related risk criteria.
\newblock In {\em Proceedings of the 29th International Coference on International Conference on Machine Learning}, pages 1651--1658, 2012.

\bibitem{thomaz2006reinforcement}
Andrea~L Thomaz and Cynthia Breazeal.
\newblock Reinforcement learning with human teachers: evidence of feedback and guidance with implications for learning performance.
\newblock In {\em Proceedings of the 21st national conference on Artificial intelligence-Volume 1}, pages 1000--1005, 2006.

\bibitem{turchetta2021safety}
Matteo Turchetta.
\newblock {\em Safety and Robustness in Reinforcement Learning}.
\newblock PhD thesis, ETH Zurich, 2021.

\bibitem{turchetta2016safe}
Matteo Turchetta, Felix Berkenkamp, and Andreas Krause.
\newblock Safe exploration in finite markov decision processes with gaussian processes.
\newblock {\em Advances in Neural Information Processing Systems}, 29, 2016.

\bibitem{turchetta2019safe}
Matteo Turchetta, Felix Berkenkamp, and Andreas Krause.
\newblock Safe exploration for interactive machine learning.
\newblock {\em Advances in Neural Information Processing Systems}, 32, 2019.

\bibitem{van2018contrastive}
J~van~der Waa, J~van Diggelen, K~van~den Bosch, and M~Neerincx.
\newblock Contrastive explanations for reinforcement learning in terms of expected consequences.
\newblock In {\em Proceedings of the Workshop on Explainable AI on the IJCAI conference, Stockholm, Sweden.}, volume~37, 2018.

\bibitem{wachi2018safe}
Akifumi Wachi, Yanan Sui, Yisong Yue, and Masahiro Ono.
\newblock Safe exploration and optimization of constrained mdps using gaussian processes.
\newblock In {\em Proceedings of the AAAI Conference on Artificial Intelligence}, volume~32, 2018.

\bibitem{xiong2021safety}
Hao Xiong and Xiumin Diao.
\newblock Safety robustness of reinforcement learning policies: A view from robust control.
\newblock {\em Neurocomputing}, 422:12--21, 2021.

\bibitem{yu2022safe}
Dongjie Yu, Wenjun Zou, Yujie Yang, Haitong Ma, Shengbo~Eben Li, Jingliang Duan, and Jianyu Chen.
\newblock Safe model-based reinforcement learning with an uncertainty-aware reachability certificate.
\newblock {\em arXiv preprint arXiv:2210.07553}, 2022.

\bibitem{yuan2022situ}
Luyao Yuan, Xiaofeng Gao, Zilong Zheng, Mark Edmonds, Ying~Nian Wu, Federico Rossano, Hongjing Lu, Yixin Zhu, and Song-Chun Zhu.
\newblock In situ bidirectional human-robot value alignment.
\newblock {\em Science robotics}, 7(68):eabm4183, 2022.

\bibitem{zhao2021dear}
Xiangyu Zhao, Changsheng Gu, Haoshenglun Zhang, Xiwang Yang, Xiaobing Liu, Hui Liu, and Jiliang Tang.
\newblock Dear: Deep reinforcement learning for online advertising impression in recommender systems.
\newblock In {\em Proceedings of the AAAI Conference on Artificial Intelligence}, volume~35, pages 750--758, 2021.

\end{thebibliography}
\bibliographystyle{plain}

\end{document}